\documentclass[letterpaper,10pt,conference]{ieeeconf}
\IEEEoverridecommandlockouts
\usepackage{graphicx}
\usepackage{amsmath,amssymb}
\usepackage{booktabs}
\usepackage{array}
\usepackage{url}
\usepackage{placeins}
\usepackage{float}
\usepackage{xcolor}
\usepackage{eurosym}
\definecolor{darkblue}{RGB}{0,0,139}
\usepackage[colorlinks=true, linkcolor=black, citecolor=black, urlcolor=black]{hyperref}
\renewcommand{\footnoterule}{%
  \kern-3pt
  \hrule width 0.4\columnwidth
  \kern 2.6pt
}
\title{\LARGE \bf
Human-Like Redundancy Resolution for Real-Time Bimanual Humanoid Teleoperation}

\author{\begin{tabular}{c}
Chao Tan, Ming Shin Cindy Sung, Shuxian Liang$^{\dagger}$, Chaoqun Wan, Zhen Huang, Xian-Sheng Hua$^{\dagger}$\\
Institute of AI for Engineering, Tongji University
\end{tabular}%
\thanks{$^{\dagger}$Corresponding authors.}}

\begin{document}
\maketitle
\thispagestyle{empty}
\pagestyle{empty}

\begin{abstract}
Real-time bimanual humanoid teleoperation requires accurate wrist tracking while keeping arm configurations predictable. Because a 7-DoF arm admits multiple elbow configurations for the same wrist pose, wrist-only tracking can cause elbow drift or abrupt configuration changes. Fixed anthropomorphic constraints can stabilize elbow posture, but they can also block the redundant adjustments needed near singular configurations. We propose a human-like redundancy-resolution framework that combines adaptive elbow guidance with singularity-robust wrist tracking. Shoulder--wrist geometric retargeting generates robot-scale wrist targets. An adaptive elbow-offset criterion uses the offset direction to guide posture and scales this guidance by the offset magnitude, so the redundant degree of freedom is progressively released as the human arm straightens. A hierarchical velocity solver prioritizes wrist tracking and projects the adaptive elbow guidance into the wrist-task null space. In a fixed-wrist-target ablation, adaptive guidance reduces mean elbow-direction error by 54.4\% compared with removing this guidance. In a workspace-boundary stress test, singularity-robust tracking reduces peak raw joint-velocity demand by 97.9\% compared with the undamped pseudoinverse. Real-world bimanual tasks and cross-platform retargeting show applicability across robot morphologies.

\noindent\textit{Project website:} \href{https://human-like-redundancy.github.io/}{\textcolor{darkblue}{https://human-like-redundancy.github.io/}}
\end{abstract}

\section{Introduction}
Real-time bimanual humanoid teleoperation transfers human motion to robots for dexterous manipulation and demonstration collection [1]--[4]. Accurate wrist tracking preserves the operator's end-effector intent, while predictable arm configurations and smooth motion support effective execution. For a 7-degree-of-freedom arm, a six-dimensional wrist task generally leaves one redundant degree of freedom; resolving this redundancy selects the elbow configuration without changing the commanded wrist pose.

\begin{figure}[t]
  \centering
  \includegraphics[width=\columnwidth]{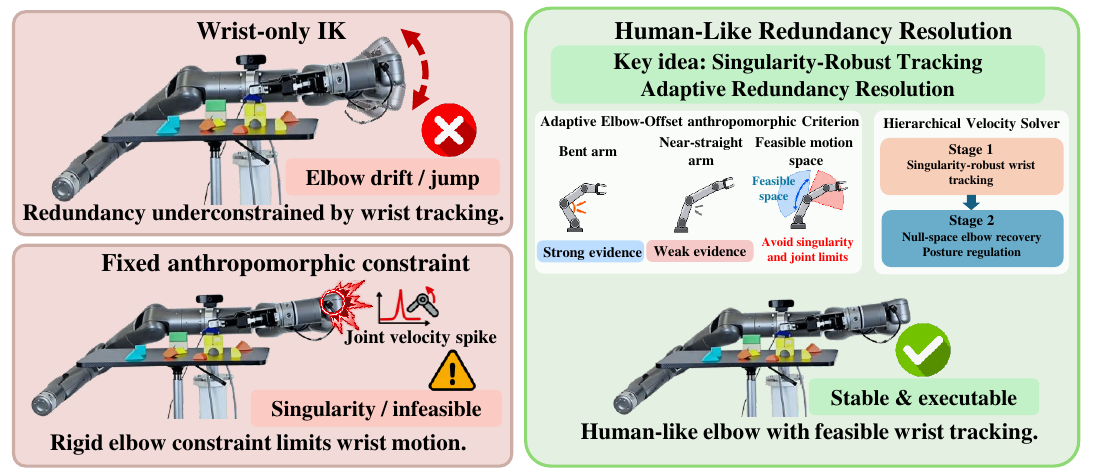}
  \caption{Human-like redundancy resolution. Wrist-only inverse kinematics leaves the elbow underconstrained, whereas a fixed anthropomorphic objective can become unreliable near full extension. Our method adapts elbow evidence and combines it with singularity-robust wrist tracking.}
  \label{fig:teaser}
\end{figure}

Wrist-only inverse kinematics leaves this redundant degree of freedom unconstrained, which can cause elbow drift or abrupt configuration changes between nearby wrist targets (the top left part of Fig.~\ref{fig:teaser}). Anthropomorphic retargeting instead uses human keypoints, geometric redundancy parameters, or posture objectives to improve human--robot configuration consistency [5], [6], [10], [12], [14]. However, enforcing an anthropomorphic target rigidly can block the redundant adjustment needed near a robot singularity, degrading wrist tracking or making the objectives infeasible (the bottom left part of Fig.~\ref{fig:teaser}). The strength of anthropomorphic guidance should also reflect the geometric reliability of the observed human posture.

Two distinct failure modes must therefore be addressed. First, when the human shoulder, elbow, and wrist approach collinearity, the elbow's lateral offset from the shoulder--wrist axis approaches zero. Its normalized direction then becomes sensitive to measurement noise, so strongly enforcing that direction can produce unnecessary posture corrections from weak geometric evidence. Second, when the robot wrist Jacobian is poorly conditioned, modest task-space commands can produce large joint-velocity requests. Weakening unreliable elbow guidance addresses the first failure mode by releasing the redundant degree of freedom, but it does not regularize the primary wrist task. Conversely, singularity-robust wrist execution limits velocity amplification but does not select a human-like elbow posture.

We therefore propose a human-like redundancy-resolution framework that combines geometry-adaptive elbow guidance with singularity-robust wrist tracking (the right part of Fig.~\ref{fig:teaser}). Shoulder--wrist geometric retargeting first generates robot-scale wrist targets. An adaptive elbow-offset anthropomorphic criterion uses the offset direction to specify the desired posture and the offset magnitude to measure the reliability of that direction. As the human arm approaches full extension, the criterion progressively weakens and then deactivates elbow guidance. A hierarchical velocity solver suppresses near-singular joint-velocity amplification for the primary wrist task and applies the adaptive posture guidance in its null space. Experiments evaluate elbow-posture consistency, wrist tracking, and near-singular velocity demand, while cross-platform retargeting and real-world bimanual tasks assess practical applicability.

The main contributions are:
\begin{itemize}
\item \textbf{Adaptive anthropomorphic redundancy resolution.} We introduce an elbow-offset criterion that uses the offset direction to guide posture and the offset magnitude to adjust the strength of that guidance. It thereby transitions smoothly from posture guidance in bent-arm configurations to redundancy release near full extension.

\item \textbf{Singularity-robust task-priority coordination.} We integrate adaptive elbow guidance with J-PARSE-based wrist tracking in a task-priority framework. The framework suppresses near-singular joint-velocity amplification for the primary wrist task and uses the resulting null space to improve arm posture.

\item \textbf{Experimental and cross-platform validation.} Adding AEAC reduces mean elbow-direction error by 54.4\% compared with removing it in fixed-wrist ablations, and singularity-robust tracking reduces peak raw joint-velocity demand by 97.9\% relative to the undamped pseudoinverse in boundary stress tests. Qualitative cross-platform retargeting and real-robot bimanual tasks show practical feasibility.

\end{itemize}

\section{Related Work}
\subsection{Bimanual Humanoid Teleoperation}
Recent bimanual teleoperation systems transfer operator hand motion to humanoid platforms via immersive visual feedback or wearable interfaces~[3], [4], with whole-body extensions coordinating torso and legs in real time~[1], [23], [24], [25] and cross-platform frameworks integrating sensing, retargeting, and execution~[17], [22], [26]. End-effector retargeting preserves task-space intent but leaves the elbow in the robot null space~[13], [18], while keypoint-based methods add shoulder and elbow landmarks for richer posture information~[7], [14] yet degrade near full extension or self-occlusion. Our method uses the same human keypoints for redundancy guidance, but adapts their influence according to the geometric observability of the elbow and preserves wrist tracking through task-priority control.

\subsection{Anthropomorphic Retargeting}
Geometric redundancy parameterizations~[5], [8], closed-form solvers~[6], and posture-constrained mappings~[9], [10] address the elbow redundancy that remains after the wrist pose is fixed. Other approaches encode anthropomorphic behavior through monocular arm imitation~[11], [12], hybrid redundancy resolution~[19], or object-aware posture selection~[20], [21]. Arm-plane-normal matching~\cite{ref10} represents anthropomorphic posture through a unit direction. Our AEAC couples directional alignment with the elbow's perpendicular offset magnitude, making guidance strength dependent on human arm geometry. This extends direction-only matching into a geometry-adaptive redundancy objective that regulates both posture alignment and its activation.

\subsection{Analytical Geometric Redundancy Solvers}
Manipulability measures~[27], [29], [35] and redundancy-resolution frameworks~[28], [31] quantify singularity proximity and exploit the null space for secondary objectives; for 7-DoF arms, the wrist task leaves one degree of freedom that can be controlled through null-space motion~[5], [6]. Singularity-robust damping and task-priority mappings~[8], [13], [30] improve numerical behavior near poorly conditioned configurations, while hierarchical formulations~[33], [34] explicitly preserve primary task priorities. Our method follows this task-priority principle with a J-PARSE velocity mapping for the stacked wrist task and uses adaptive elbow guidance only in the resulting null space.

\section{Methodology}
\begin{figure*}[t]
  \centering
  \includegraphics[width=\textwidth]{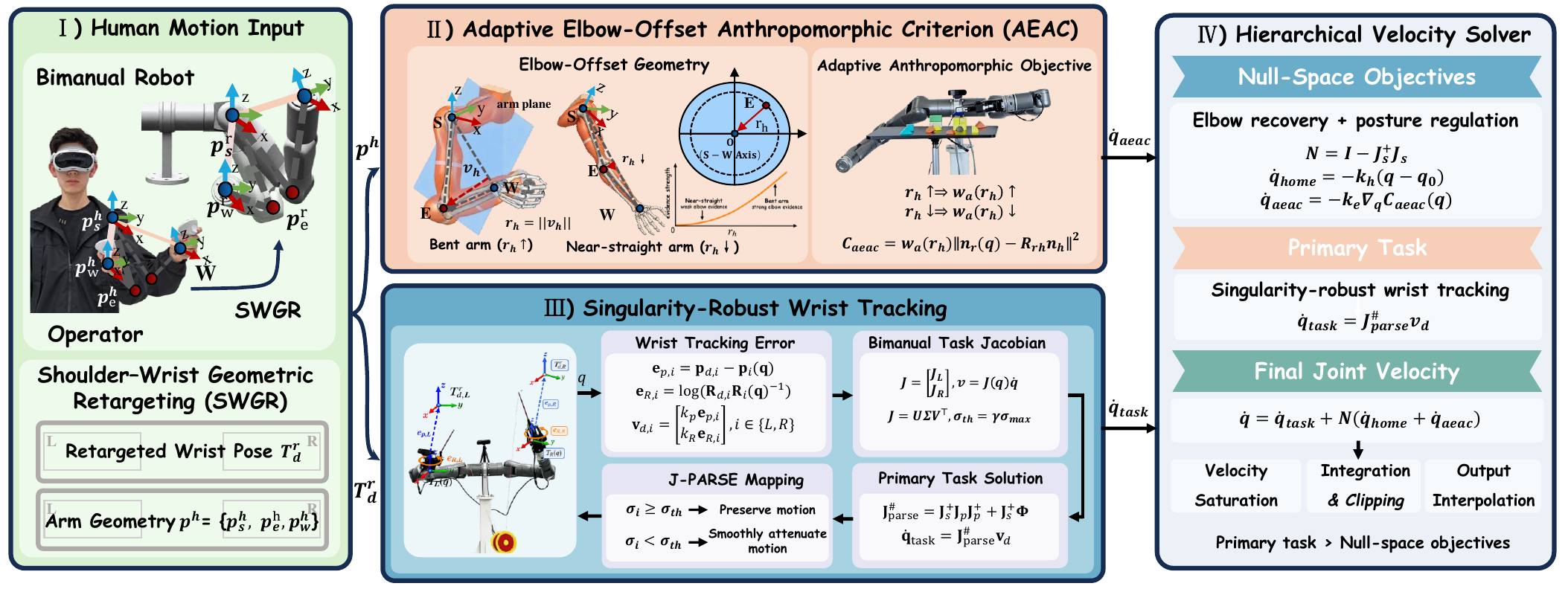}
  \caption{\textbf{Framework of Human-Like Redundancy Resolution for Real-Time Bimanual Humanoid Teleoperation.} I)~\emph{Shoulder--Wrist Geometric Retargeting (SWGR)}: XR shoulder, elbow, and wrist keypoints with controller poses are retargeted to robot-scale wrist targets. II)~\emph{Adaptive Elbow-Offset Anthropomorphic Criterion (AEAC)}: the elbow-offset direction guides redundancy with adaptive weighting. III)~\emph{Singularity-Robust Wrist Tracking}: J-PARSE regularizes the stacked wrist Jacobian. IV)~\emph{Hierarchical Velocity Solver}: the primary wrist task is combined with AEAC and posture regulation in the task null space to generate joint commands.}
  \label{fig:main}
\end{figure*}

We develop a real-time motion-retargeting and hierarchical velocity-solving framework for bimanual teleoperation with two 7-DoF humanoid arms. Given XR shoulder, elbow, and wrist keypoints together with controller poses, the system generates robot wrist targets through geometric retargeting, resolves arm redundancy with an adaptive elbow-offset anthropomorphic criterion, and computes joint commands using a hierarchical velocity solver with singularity-robust wrist tracking. As shown in Fig.~\ref{fig:main}, the method comprises four components: (i)~Shoulder--Wrist Geometric Retargeting (SWGR, Section~III-B), (ii)~the adaptive elbow-offset anthropomorphic criterion (AEAC, Section~III-C), (iii)~singularity-robust wrist tracking (Section~III-D), and (iv)~the hierarchical velocity solver (Section~III-E).

\subsection{Problem Formulation}
\label{sec:problem}
We consider a humanoid robot with two 7-DoF arms, whose joint configuration is denoted by $\mathbf q=[\mathbf q_L^{\mathsf T}\ \mathbf q_R^{\mathsf T}]^{\mathsf T}\in\mathbb R^{14}$, with $\mathbf q_i\in\mathbb R^7$. At time $t$, the XR interface provides
\begin{equation}
\mathcal X_t=
\left\{
\mathbf p_{s,i}^{h},\mathbf p_{e,i}^{h},\mathbf p_{w,i}^{h},
\mathbf R_{c,i}^{h}
\right\}_{i\in\{L,R\}},
\label{eq:xr-input}
\end{equation}
where $\mathbf p_{s,i}^{h}$, $\mathbf p_{e,i}^{h}$, and $\mathbf p_{w,i}^{h}$ are human shoulder, elbow, and wrist positions, and $\mathbf R_{c,i}^{h}$ is the controller orientation. Given the retargeted wrist velocity $\mathbf v_d$, we compute a singularity-robust primary wrist velocity
\begin{equation}
\dot{\mathbf q}_{\mathrm{task}}
=
\mathbf J_{\mathrm{JPARSE}}^{\#}\mathbf v_d,
\end{equation}
and project secondary posture commands into the null space of the safety Jacobian:
\begin{equation}
\dot{\mathbf q}
=
\dot{\mathbf q}_{\mathrm{task}}
+
\mathbf N\dot{\mathbf q}_{\mathrm{sec}},
\end{equation}
where $\mathbf N$ is the J-PARSE task null-space projector and $\dot{\mathbf q}_{\mathrm{sec}}$ contains AEAC and posture-regulation terms. With solver period $\Delta t$, the state update is
\begin{equation}
\mathbf q_{t+1}=\operatorname{clip}\!\left(
\mathbf q_t+\dot{\mathbf q}_t\Delta t,\,
\mathbf q_{\min},\mathbf q_{\max}\right).
\end{equation}

\subsection{Shoulder--Wrist Geometric Retargeting}
\label{sec:retargeting}
We use Shoulder--Wrist Geometric Retargeting (SWGR, Fig.~\ref{fig:main}(I)) to convert human upper-arm motion into robot-scale wrist targets. Let $\mathbf p_s^h$ and $\mathbf p_w^h$ denote the human shoulder and wrist positions, $\mathbf p_s^r$ the corresponding robot shoulder anchor, and $\mathbf R_{rh}$ the rotation from the human tracking frame to the robot frame. The desired wrist position is
\begin{equation}
\mathbf p_{w,d}^{r}
=
\mathbf p_s^r
+\alpha\mathbf R_{rh}
\left(\mathbf p_w^h-\mathbf p_s^h\right),
\label{eq:swgr-position}
\end{equation}
where $\alpha=l_r/l_h$, and $l_h$ and $l_r$ are the human and robot reference shoulder--wrist lengths. Because only a relative shoulder--wrist vector is used, translations of the XR world origin do not affect the retargeted position.

The controller orientation is mapped using a calibrated tool offset according to $\mathbf R_{d}^{r}=\mathbf R_{rh}\mathbf R_{c}^{h}\mathbf R_{\mathrm{off}}$.
The desired wrist pose is therefore
\begin{equation}
\mathbf T_d=
\begin{bmatrix}
\mathbf R_d^r & \mathbf p_{w,d}^{r}\\
\mathbf 0^{\mathsf T} & 1
\end{bmatrix}.
\end{equation}

Before velocity solving, each target position is smoothed by a One Euro filter~\cite{ref39}. The target orientation is filtered by quaternion spherical interpolation~\cite{ref40}, which avoids directly averaging rotation matrices. These filters act on the desired pose only; they do not alter the task hierarchy or the null-space computation.

\subsection{Adaptive Elbow-Offset Anthropomorphic Criterion}
\label{sec:elbow}
The 7-DoF wrist task leaves one redundant degree of freedom per arm. The Adaptive Elbow-Offset Anthropomorphic Criterion (AEAC, Fig.~\ref{fig:main}(II)) uses the human elbow observation to guide this redundancy while reducing its influence when the elbow direction is poorly observable. Fixed-weight constraints~\cite{ref8,ref10} can amplify tracking noise near full extension, where the shoulder, elbow, and wrist become nearly collinear. AEAC separates the elbow evidence into a direction and a magnitude: the direction defines the anthropomorphic objective, and the magnitude determines how strongly that objective is used.

\medskip
\noindent\textbf{Anthropomorphic criterion.}
For one arm, let $\mathbf p_s^h$, $\mathbf p_e^h$, and $\mathbf p_w^h$ denote the human shoulder, elbow, and wrist positions in a common frame. The shoulder--wrist axis direction is
\begin{equation}
\mathbf d_h
=
\frac{\mathbf p_w^h-\mathbf p_s^h}
{\max(\|\mathbf p_w^h-\mathbf p_s^h\|_2,\varepsilon)} .
\end{equation}
The projection of the elbow onto this axis and the perpendicular offset are
\begin{equation}
\mathbf p_{\mathrm{proj}}^h
=
\mathbf p_s^h+
\mathbf d_h\mathbf d_h^{\mathsf T}
\left(\mathbf p_e^h-\mathbf p_s^h\right).
\end{equation}
The perpendicular elbow offset is
\begin{equation}
\mathbf v_h=\mathbf p_e^h-\mathbf p_{\mathrm{proj}}^h .
\end{equation}
The human elbow direction is
\begin{equation}
\mathbf n_h
=
\frac{\mathbf v_h}{\max(\|\mathbf v_h\|_2,\varepsilon)},
\end{equation}
and its offset magnitude is denoted in the following by $h=\|\mathbf v_h\|_2$.
The robot direction $\mathbf n_r(\mathbf q)$ is computed by applying the same construction to the forward-kinematic positions $\mathbf p_s^r(\mathbf q)$, $\mathbf p_e^r(\mathbf q)$, and $\mathbf p_w^r(\mathbf q)$. After transforming the human direction into the robot frame, the directional anthropomorphic loss is
\begin{equation}
\mathcal L_a(\mathbf q)
=
\left\|\mathbf n_r(\mathbf q)-\mathbf R_{rh}\mathbf n_h\right\|_2^2 .
\label{eq:aeac-direction}
\end{equation}

\medskip
\noindent\textbf{Adaptive implementation.}
The length $h$ measures how far the elbow lies from the shoulder--wrist axis and therefore how reliable the elbow direction is. As the arm straightens, $h\rightarrow0$ and $\mathbf n_h$ becomes noise-sensitive. With a deadband $\delta$ and maximum weight $w_{\max}$, the adaptive weight is
\begin{equation}
\omega_a
=
w_{\max}\operatorname{clip}
\left(
\frac{h-\delta}{l_h-\delta},0,1
\right).
\label{eq:aeac-weight}
\end{equation}
The final AEAC cost is
\begin{equation}
C_{\mathrm{AEAC}}(\mathbf q)
=
\omega_a\mathcal L_a(\mathbf q).
\label{eq:aeac-cost}
\end{equation}
When $h>\varepsilon$, differentiating the normalized direction gives
\begin{equation}
\frac{\partial \mathbf n_h}{\partial \mathbf v_h}
=
\frac{\mathbf I-\mathbf n_h\mathbf n_h^{\mathsf T}}{h},
\end{equation}
so the sensitivity of the normalized elbow direction scales inversely with the offset magnitude. AEAC therefore uses this magnitude to regulate the influence of directional alignment, retaining posture guidance when the geometry is informative and deactivating it within the extension deadband. When $h\leq\delta$, $\omega_a=0$ and the elbow observation is ignored; when the arm has a measurable lateral offset, the criterion is gradually activated. In the implementation, the gradient of Eq.~\eqref{eq:aeac-cost} is evaluated by automatic differentiation and used as a null-space velocity. AEAC is therefore a soft secondary preference, rather than a hard constraint that can override wrist tracking.

\subsection{Singularity-Robust Wrist Tracking}
\label{sec:singularity}
Motivated by J-PARSE~\cite{ref36}, which adapts task-space commands to the available mobility near singularities, we combine singularity-robust wrist tracking (Fig.~\ref{fig:main}(III)) with AEAC-based anthropomorphic redundancy resolution. For each wrist, the geometric Jacobian maps joint velocity to a six-dimensional twist; the two Jacobians are stacked as
\begin{equation}
\mathbf J(\mathbf q)
=
\begin{bmatrix}
\mathbf J_L(\mathbf q)\\
\mathbf J_R(\mathbf q)
\end{bmatrix},
\qquad
\mathbf v=\mathbf J(\mathbf q)\dot{\mathbf q}.
\end{equation}
For arm $i$, define the position and orientation errors as
\begin{equation}
\mathbf e_{p,i}=\mathbf p_{d,i}-\mathbf p_i(\mathbf q),
\qquad
\mathbf e_{R,i}
=
\operatorname{Log}\!\left(
\mathbf R_{d,i}\mathbf R_i(\mathbf q)^{-1}
\right).
\end{equation}
The desired stacked twist is
\begin{equation}
\mathbf v_d
=
\begin{bmatrix}
k_p\mathbf e_{p,L}^{\mathsf T} &
k_R\mathbf e_{R,L}^{\mathsf T} &
k_p\mathbf e_{p,R}^{\mathsf T} &
k_R\mathbf e_{R,R}^{\mathsf T}
\end{bmatrix}^{\mathsf T}.
\label{eq:desired-twist}
\end{equation}

\medskip
The J-PARSE construction is defined as follows.
We compute the singular value decomposition
\begin{equation}
\mathbf J=\mathbf U\mathbf\Sigma\mathbf V^{\mathsf T}.
\end{equation}
Let $\sigma_{\max}$ denote the largest singular value and set
\begin{equation}
\sigma_{\mathrm{th}}=\gamma\sigma_{\max}.
\end{equation}
Following J-PARSE~\cite{ref36}, let $\mathbf u_j$ and $\mathbf v_j$ denote the corresponding left and right singular vectors of the stacked wrist Jacobian. The safety Jacobian is
\begin{equation}
\mathbf J_s
=
\sum_{j=1}^{12}
\max(\sigma_j,\sigma_{\mathrm{th}})
\mathbf u_j\mathbf v_j^{\mathsf T}.
\label{eq:safety-jacobian}
\end{equation}
The inverse gain for each singular direction is
\begin{equation}
g_j=
\begin{cases}
1/\sigma_j, & \sigma_j>\sigma_{\mathrm{th}},\\[1mm]
\sigma_j/\sigma_{\mathrm{th}}^2,
& \sigma_j\leq\sigma_{\mathrm{th}}.
\end{cases}
\label{eq:jparse-gain}
\end{equation}
The singularity-robust inverse is therefore
\begin{equation}
\mathbf J^\#_{\mathrm{JPARSE}}
=
\sum_{j=1}^{12}
g_j\mathbf v_j\mathbf u_j^{\mathsf T}.
\label{eq:jparse-inverse}
\end{equation}
This mapping preserves the inverse gains of well-conditioned directions and attenuates poorly conditioned ones.
The primary wrist-task velocity is then
\begin{equation}
\dot{\mathbf q}_{\mathrm{task}}
=
\mathbf J_{\mathrm{JPARSE}}^{\#}\mathbf v_d.
\label{eq:jparse-task}
\end{equation}

\subsection{Hierarchical Velocity Solver}
\label{sec:solver}
The primary wrist task is combined with secondary objectives only after the J-PARSE mapping has been computed (Fig.~\ref{fig:main}(IV)). The associated null-space projector is
\begin{equation}
\mathbf N
=
\mathbf I-\mathbf J_s^{+}\mathbf J_s .
\label{eq:nullspace}
\end{equation}
The joint-centering velocity is
\begin{equation}
\dot{\mathbf q}_{\mathrm{home}}
=
-k_h(\mathbf q-\mathbf q_0),
\end{equation}
where $\mathbf q_0$ is the reference configuration. The AEAC velocity is obtained from the negative gradient of the adaptive anthropomorphic cost:
\begin{equation}
\dot{\mathbf q}_{\mathrm{aeac}}
=
-k_e\nabla_{\mathbf q}C_{\mathrm{AEAC}}(\mathbf q).
\label{eq:aeac-gradient}
\end{equation}
The final command before limiting is
\begin{equation}
\dot{\mathbf q}^{\ast}
=
\dot{\mathbf q}_{\mathrm{task}}
+
\mathbf N\left(
\dot{\mathbf q}_{\mathrm{home}}
+
\dot{\mathbf q}_{\mathrm{aeac}}
\right).
\label{eq:hierarchical-velocity}
\end{equation}
The projected AEAC and posture terms reshape the arm configuration within the wrist-task null space, preserving the primary task at the differential level before command limiting. The combined velocity is then scaled to satisfy the joint-speed bound, integrated, and clipped to the joint limits.

The command is bounded by the configured maximum joint speed:
\begin{equation}
\dot{\mathbf q}
=
\dot{\mathbf q}^{\ast}
\min\left(
1,
\frac{\dot q_{\max}}
{\|\dot{\mathbf q}^{\ast}\|_{\infty}+\varepsilon}
\right).
\label{eq:velocity-limit}
\end{equation}
It is then integrated and clipped to the robot joint limits using the configuration update in Section~\ref{sec:problem}. The default controller runs the velocity solver loop at $90$\,Hz with $\dot q_{\max}=3.14$\,rad/s. The position filter uses $\text{min\_cutoff}=5.0$ and $\beta=10.0$, while the orientation filter uses $\alpha=0.9$. Finally, the integrated joint targets are linearly interpolated at the $200$\,Hz output layer to improve command continuity without feeding interpolated values back into the solver state.

\begin{figure}[t]
  \centering
  \includegraphics[width=\columnwidth]{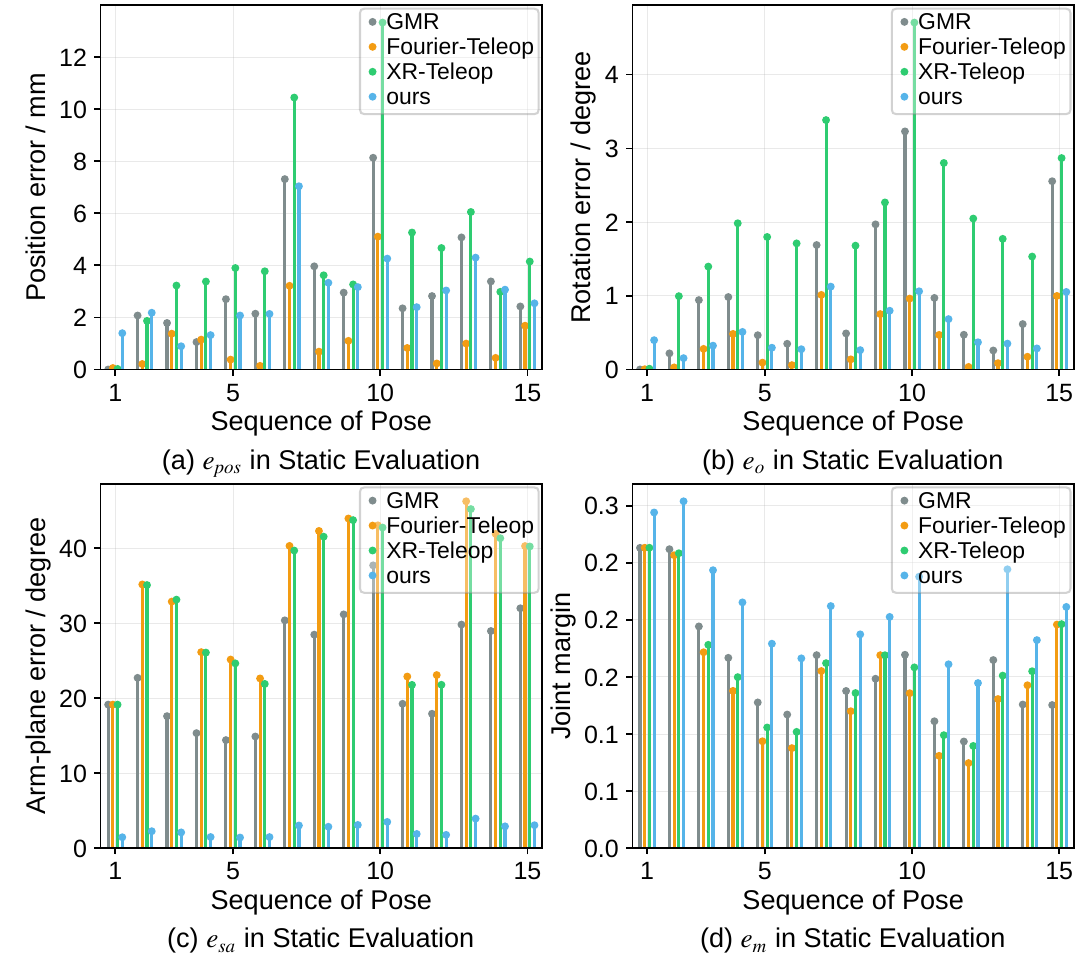}
  \caption{Static evaluation of GMR, Fourier-Teleop, XR-Teleop, and Ours over 15 poses. (a) Wrist position error $e_{\mathrm{pos}}$; (b) wrist rotation error $e_o$; (c) arm-plane error $e_{\mathrm{sa}}$; and (d) joint margin $e_m$.}
  \label{fig:static}
\end{figure}

\section{Validation and Discussion}
\label{sec:validation}

\subsection{Experimental Setup and Evaluation Metrics}
\textbf{Compared methods.} We compare three representative baselines: XR-Teleop~\cite{ref15} for wrist-centric IK, Fourier-Teleop~\cite{ref16} for posture-regularized differential IK, and GMR~\cite{ref14} for body-geometry-based retargeting. These baselines cover end-effector tracking, reference-posture regulation, and human-motion consistency. The comparison evaluates the benefits of geometry-adaptive elbow guidance for human-like redundancy resolution. We evaluate on 15 static key poses and three real-time motion sequences: elbow motion for anthropomorphic fidelity, circle motion for general motion range, and forward extension for extreme singular postures. Component ablations further isolate the contributions of AEAC and the singularity-robust mapping.

\textbf{Experimental platform.} Experiments run on a PICO~4~Ultra~\cite{ref37} or Vive Tracker~\cite{ref38} with a JAKA~K1 humanoid dual-arm robot at 90\,Hz (Ubuntu~24.04.4~LTS, Intel~i7-11800H, NVIDIA~RTX~3050 Laptop~GPU). The per-frame solve time averages 4.35\,ms, within the 11.1\,ms control budget.

\textbf{Evaluation metrics.} Functional tracking is measured by wrist position error $e_{\mathrm{pos}}=\|\mathbf p^{\mathrm{des}}-\mathbf p^{\mathrm{cur}}\|_2$ and rotation error $e_o=\|\operatorname{Log}((\mathbf R^{\mathrm{des}})^{\mathsf T}\mathbf R^{\mathrm{cur}})\|_2$ [3], [14]. Anthropomorphic posture is measured by arm-plane error $e_{\mathrm{sa}}=|\operatorname{atan2}(\sin(\psi^{h}-\psi^{r}),\cos(\psi^{h}-\psi^{r}))|$ [5], [6], [8], where $\psi^{h}$ and $\psi^{r}$ are signed swivel angles in the plane orthogonal to the shoulder--wrist axis; the time-series plots label the same metric as swivel-angle error. Joint safety is reported by the joint margin $e_m$ [31], [32], and singularity behavior is evaluated separately [27], [30], [36].

\subsection{Static Performance}

Fig.~\ref{fig:static} summarizes the static-pose results. The proposed method achieves the lowest arm-plane error across the 15 static poses, reaching $2.42\pm0.83$\textdegree{}, compared with $23.99\pm7.55$\textdegree{} for GMR. This improvement in human--robot posture consistency is achieved with a wrist position error of $2.87\pm1.52$ mm and a rotation error of $0.53\pm0.33$\textdegree{}.

\subsection{Dynamic Performance}

\begin{figure}[t]
    \centering
    \includegraphics[width=\columnwidth]{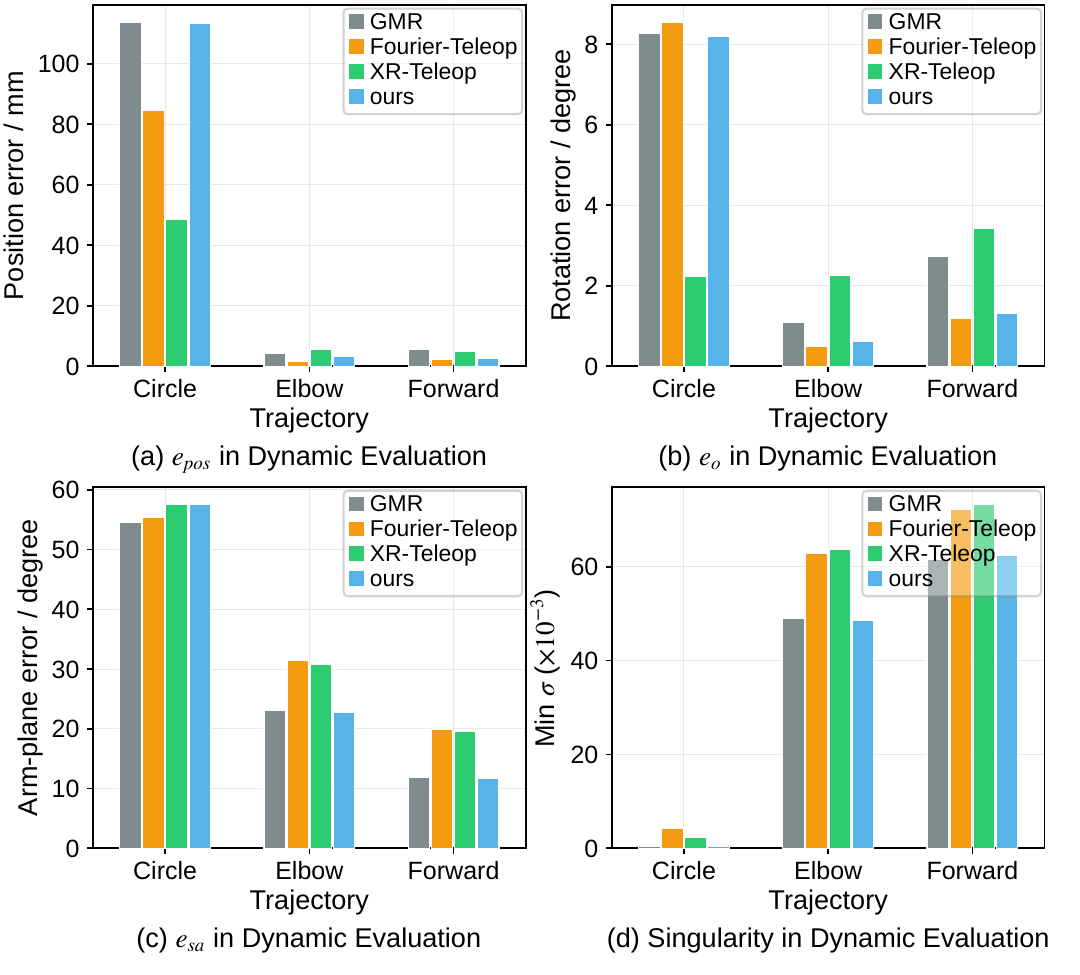}
    \caption{Dynamic tracking comparison of GMR, Fourier-Teleop, XR-Teleop, and Ours on elbow, circle, and forward-extension trajectories. (a)~Wrist position error $e_{\mathrm{pos}}$; (b)~wrist rotation error $e_o$; (c)~arm-plane error $e_{\mathrm{sa}}$; (d)~singularity margin.}
    \label{fig:dynamic}
\end{figure}

\noindent\textbf{Overall dynamic performance.}
Fig.~\ref{fig:dynamic} compares three trajectories: elbow motion, circle motion, and forward extension. The circle case is a limitation: XR-Teleop gives the lowest wrist errors (48.5\,mm, 2.22\textdegree{}), while Ours gives 113.4\,mm and 8.19\textdegree{} and does not improve arm-plane error (57.5\textdegree{}). On the elbow and forward-extension trajectories, however, Ours obtains the lowest arm-plane errors, 22.7\textdegree{} and 11.7\textdegree{}, with low wrist errors. The benefit is therefore targeted redundancy resolution and singularity-sensitive behavior, not uniformly best dynamic tracking.

\begin{figure}[!t]
    \centering
    \includegraphics[width=\columnwidth]{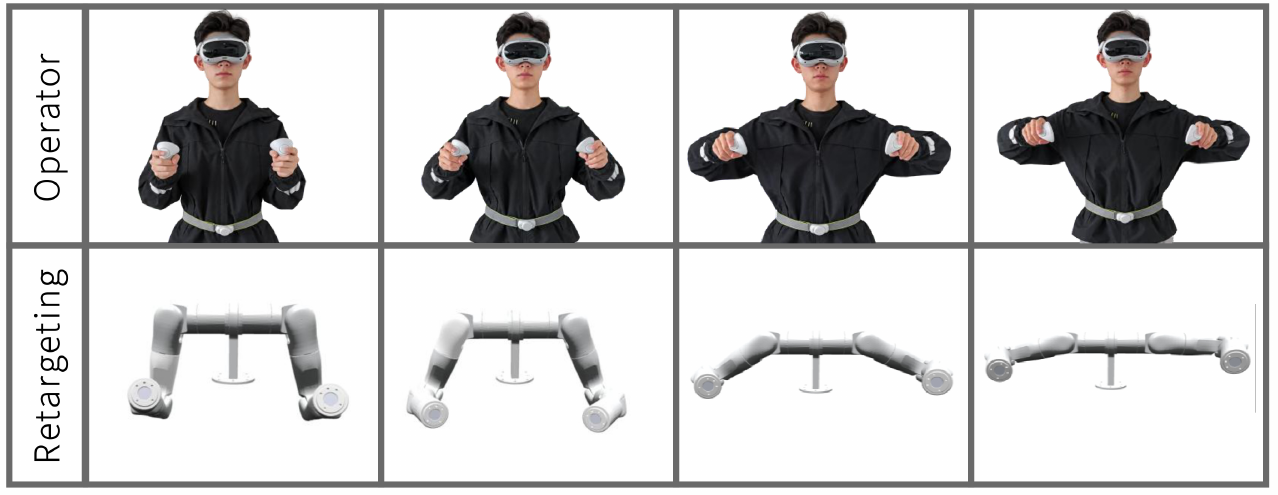}
    {\footnotesize (a) Representative dynamic anthropomorphic poses\par}
    \vspace{1mm}
    \includegraphics[width=\columnwidth]{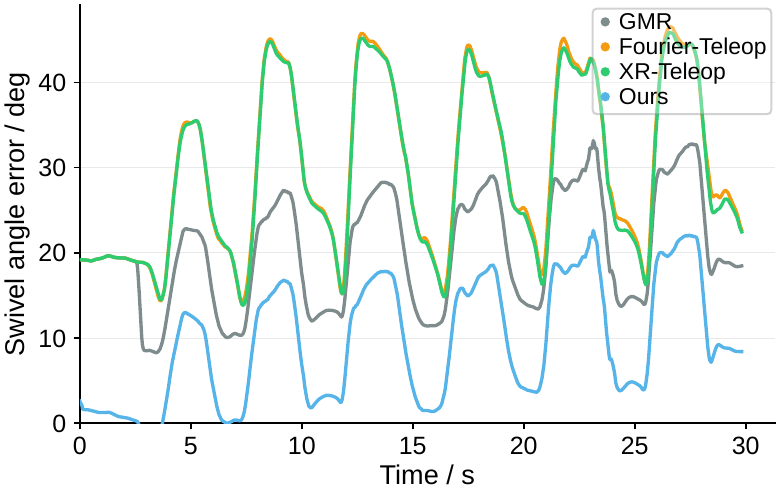}
    {\footnotesize (b) Swivel-angle error trajectories\par}
    \caption{Dynamic anthropomorphic behavior. (a)~Human and robot arm configurations at selected frames of the elbow-motion sequence. (b)~Swivel-angle error trajectories of GMR, Fourier-Teleop, XR-Teleop, and Ours.}
    \label{fig:anthro_vis}
\end{figure}

\noindent\textbf{Anthropomorphic behavior.}
Fig.~\ref{fig:anthro_vis}(a) shows representative configurations from the elbow-motion sequence, and Fig.~\ref{fig:anthro_vis}(b) plots the same arm-plane metric over time. Ours is best on elbow motion and forward extension. The circle trajectory is executed at high speed and drives the arms beyond the nominal workspace, serving primarily as a stability test rather than a posture-accuracy benchmark; the larger arm-plane error (57.52\textdegree{} versus 54.60\textdegree{} for GMR) reflects this trajectory-dependent trade-off. AEAC therefore helps when the elbow geometry is informative, but does not guarantee lower posture error for every motion pattern.

\begin{figure}[H]
    \centering
    \includegraphics[width=\columnwidth]{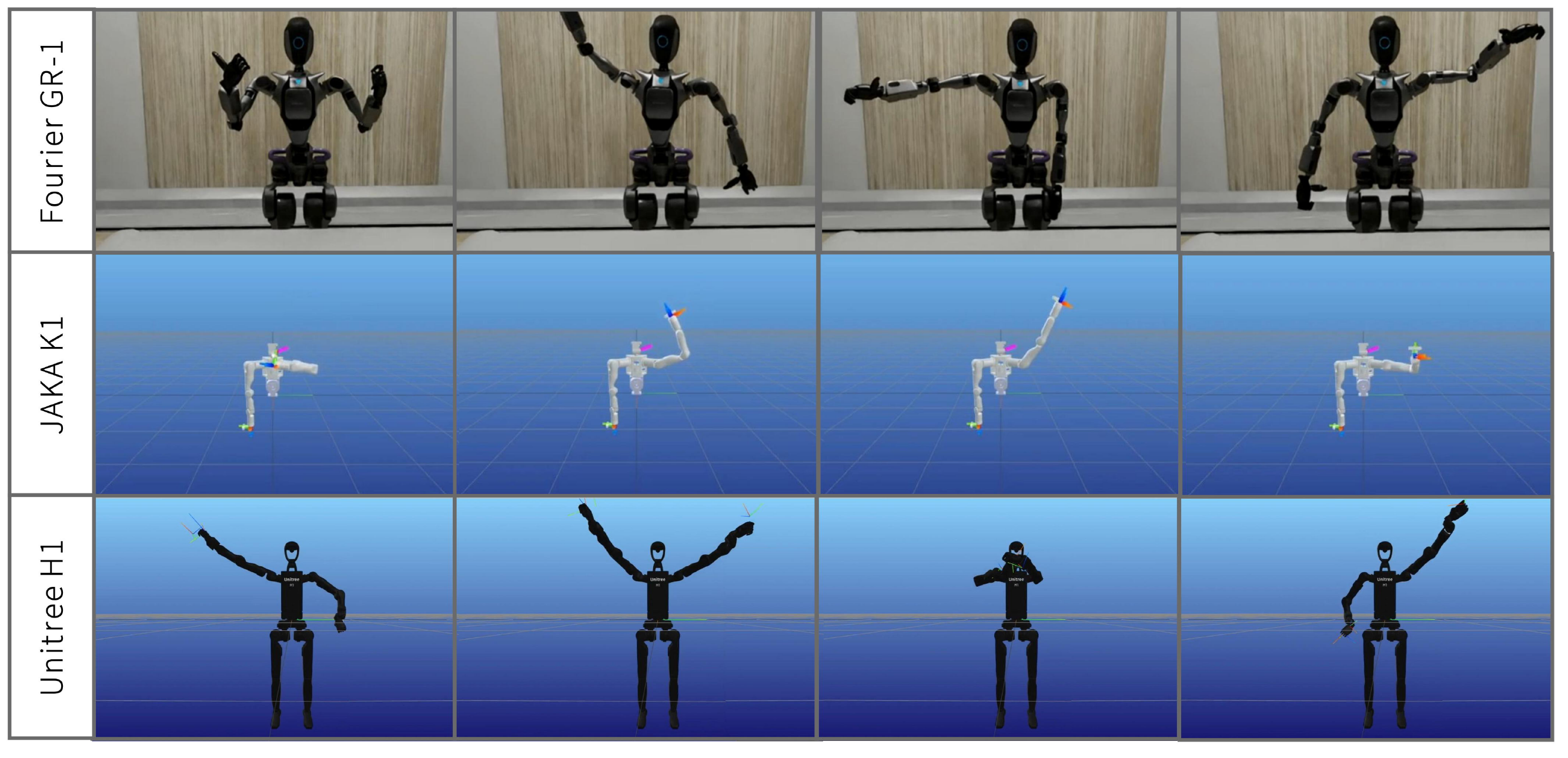}
    \caption{Cross-platform retargeting on Fourier GR-1, JAKA~K1, and Unitree H1 using identical parameters.}
    \label{fig:other}
\end{figure}

\begin{figure}[!t]
    \centering
    \includegraphics[width=\columnwidth]{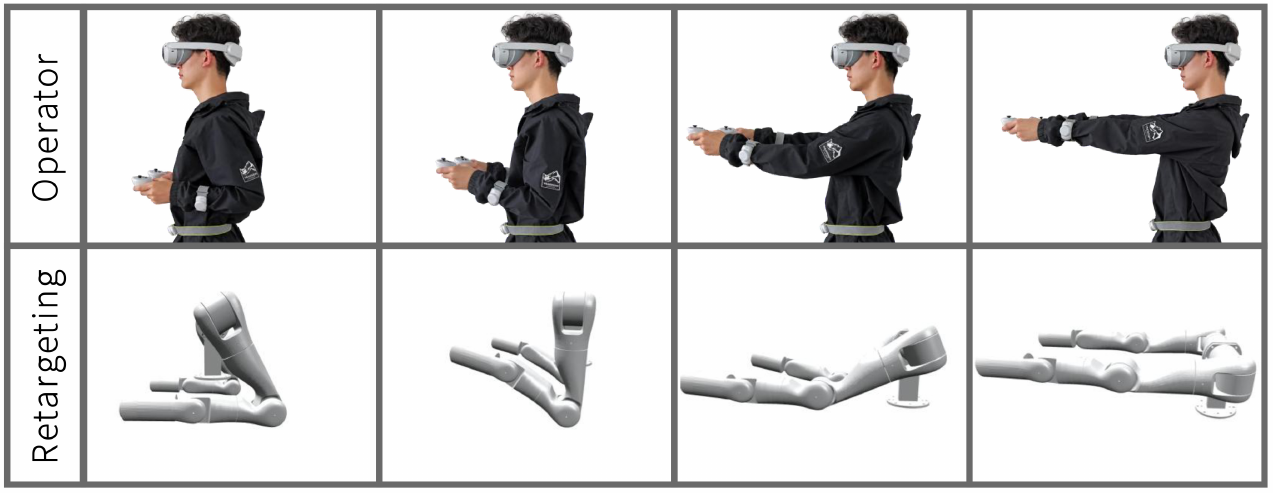}
    {\footnotesize (a) Forward-extension motion sequence\par}
    \vspace{1mm}
    \includegraphics[width=\columnwidth]{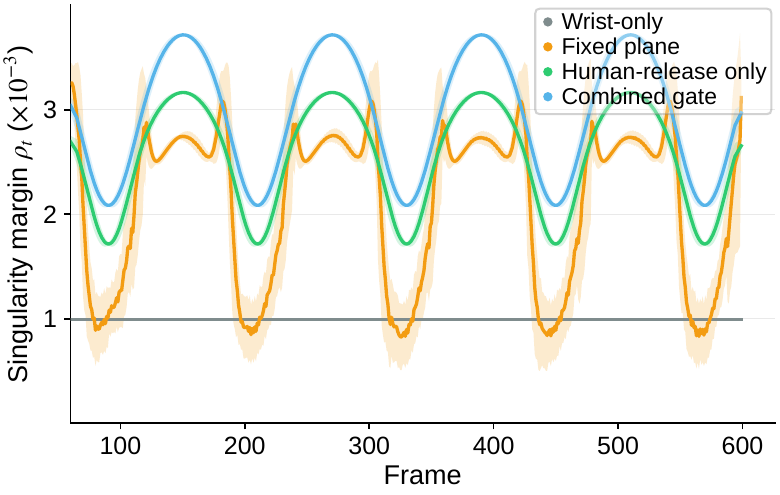}
    {\footnotesize (b) Frame-wise singularity margin\par}
    \caption{Dynamic singularity behavior. (a)~Forward-extension motion sequence and robot configurations near full extension. (b)~Frame-wise normalized singularity margin $\rho_t$ after the initial transient.}
    \label{fig:stress_vis}
\end{figure}

\noindent\textbf{Singularity behavior.}
Fig.~\ref{fig:stress_vis}(b) shows that Ours maintains a higher and more stable singularity margin than the fixed-plane and human-release-only variants during repeated extension and recovery. Fig.~\ref{fig:stress_vis}(a) illustrates the associated forward-extension motion. The complete method remains executable near the workspace boundary without the abrupt joint motion observed for a rigid anthropomorphic constraint. Detailed component-level comparisons are given in Section~\ref{sec:ablation}.

\subsection{Cross-Platform and Real-Robot Validation}

\noindent\textbf{Cross-platform validation.}
We deploy the solver on three distinct humanoid platforms without parameter retuning: Fourier GR-1 (wheeled dual-arm), JAKA~K1 (humanoid dual-arm), and Unitree H1 (full-size humanoid). Fig.~\ref{fig:other} provides qualitative cross-platform demonstrations across the different morphologies.

\begin{figure*}[!t]
    \centering
    \includegraphics[width=\textwidth]{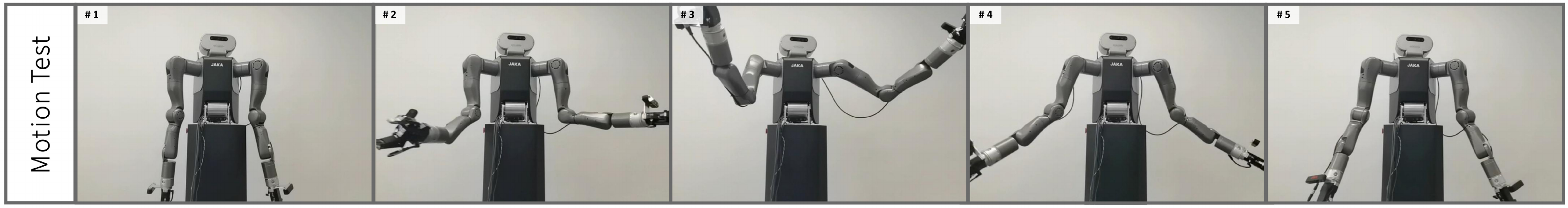}
    \parbox{\textwidth}{\footnotesize (a) The robot starts from a neutral posture (\#1 -- Neutral), performs wide arm extension (\#2 -- Extend), reaches a near-singular raised-arm posture (\#3 -- Singular), transitions through a large-range motion (\#4 -- Recover), and returns to a stable posture (\#5 -- Return).\par}
    \vspace{1mm}
    \includegraphics[width=\textwidth]{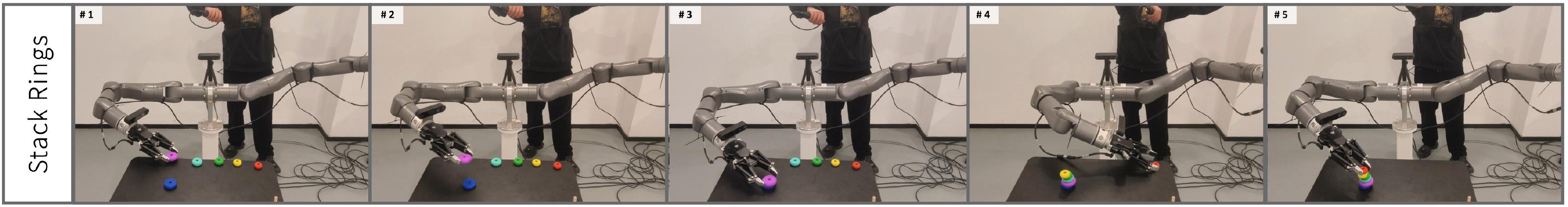}
    \parbox{\textwidth}{\footnotesize (b) The robot grasps the first ring (\#1 -- Grasp), transports it to the target (\#2 -- Transport), stacks it onto the base (\#3 -- Stack), and coordinates the other arm to pick (\#4 -- Pick) and place the next ring (\#5 -- Place).\par}
    \vspace{1mm}
    \includegraphics[width=\textwidth]{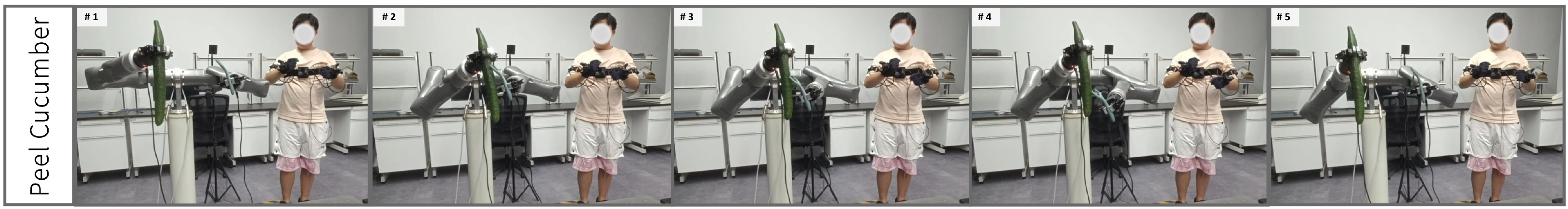}
    \parbox{\textwidth}{\footnotesize (c) The robot secures the cucumber and aligns the peeler (\#1, \#2 -- Align), performs the peeling stroke along its surface (\#3, \#4 -- Peel), and retracts the tool to complete the action (\#5 -- Retract).\par}
    \caption{Real-robot validation. (a)~Motion sequence covering neutral, extension, near-singular, recovery, and return phases. (b)~Stack Rings: bimanual grasp, transport, and placement. (c)~Peel Cucumber: one arm holds the object while the other performs a continuous tool stroke.}
    \label{fig:real}
\end{figure*}

\noindent\textbf{Real-robot task deployment.}
We further deploy the solver on one motion sequence and two bimanual tasks requiring sustained contact, tight inter-arm coordination, and operation near singularities. Fig.~\ref{fig:real} shows representative snapshots. The motion sequence covers neutral, extension, near-singular, recovery, and return phases. In \emph{Stack Rings}, both arms coordinate grasping, transport, and placement; in \emph{Peel Cucumber}, one arm holds the object while the other performs a continuous tool stroke. Real-robot demonstrations illustrate the feasibility of the framework for bimanual manipulation.

\subsection{Ablation Experiments and Analysis}
\label{sec:ablation}
\noindent\textbf{Forward-extension stress test.}
The forward-extension trajectory drives the arm through near-full-extension singular postures. We compare wrist-only retargeting, a fixed swivel target, AEAC alone, and AEAC with singularity-robust mapping. All configurations share the same input, robot model, and control frequency. Table~\ref{tab:ablation} reports arithmetic means over post-transient frames 60--599. Its singularity column uses the same normalized margin $\rho_t$ and window as Fig.~\ref{fig:stress_vis}(b), not $\min(\sigma)$ or cross-arm sums.

\begin{table}[!t]
\centering
\footnotesize
\setlength{\tabcolsep}{1.8pt}
\caption{Forward-extension ablation.}
\label{tab:ablation}
\begin{tabular}{@{}lrrrr@{}}
\toprule
\textbf{Config.} & \textbf{Post./deg}$\downarrow$ & \textbf{Pos./mm}$\downarrow$ & \textbf{$\rho_t/(10^{-3})$}$\uparrow$ & \textbf{$\bar{\ddot q}$/(rad/s$^2$)}$\downarrow$ \\
\midrule
Wrist-only           & 65.65 & 5.67  & 0.991 & \textbf{2.82} \\
Fixed swivel target   & \textbf{13.99} & 10.27 & 2.115 & 168.98 \\
Human-side adaptive  & 16.88 & \textbf{5.66} & 2.548 & 3.02 \\
Combined method      & 17.23 & \textbf{5.66} & \textbf{2.944} & 3.20 \\
\bottomrule
\end{tabular}
\end{table}

Post. is anthropomorphic posture error, Pos. is wrist position error, $\rho_t$ is normalized singularity margin, and $\bar{\ddot q}$ is mean joint acceleration; ``anthro.'' denotes anthropomorphic.

The fixed swivel target reaches a large mean acceleration, 168.98~rad/s$^2$, because it forces a constant swivel target when elbow evidence is weak. AEAC reduces this mean to 3.02~rad/s$^2$ and raises mean $\rho_t$ from $2.115\times10^{-3}$ to $2.548\times10^{-3}$. Adding singularity-robust mapping further raises mean $\rho_t$ to $2.944\times10^{-3}$ while keeping mean acceleration at 3.20~rad/s$^2$.

\begin{figure}[t]
  \centering
  \includegraphics[width=\columnwidth]{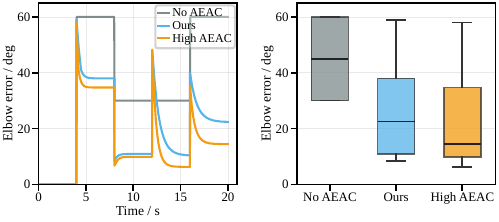}
  \caption{Anthropomorphic ablation. The proposed AEAC setting reduces elbow-direction error while avoiding the larger motion demand of high gain.}
  \label{fig:anthro_ablation}
\end{figure}

\begin{table}[t]
\centering
\footnotesize
\setlength{\tabcolsep}{2.5pt}
\caption{Anthropomorphic ablation with fixed wrist targets.}
\label{tab:anthro_ablation}
\begin{tabular}{@{}lrrrr@{}}
\toprule
\textbf{Method} & \textbf{Mean elbow}$\downarrow$ & \textbf{P90 elbow}$\downarrow$ & \textbf{Position}$\downarrow$ & \textbf{$\dot q$}$\downarrow$ \\
\midrule
Ours w/o AEAC       & 45.0 & 60.0 & 0.0  & 0.00 \\
Ours (default)      & \textbf{20.5} & 38.0 & 9.8  & 0.69 \\
Ours (high gain)    & 16.3 & \textbf{34.7} & 24.3 & 1.38 \\
\bottomrule
\end{tabular}
\end{table}

\noindent\textbf{Anthropomorphic redundancy resolution.}
We isolate AEAC by fixing the bimanual wrist targets and changing the desired elbow direction. Figure~\ref{fig:anthro_ablation} and Table~\ref{tab:anthro_ablation} show that Ours reduces the mean elbow-direction error from $45.0^\circ$ to $20.5^\circ$ while keeping the mean wrist position error at $9.8$\,mm. The higher-gain setting reduces elbow error further, but increases wrist error and raw joint velocity. Thus, the proposed setting uses the redundant degree of freedom effectively without replacing the primary wrist task.

\begin{figure}[t]
  \centering
  \includegraphics[width=\columnwidth]{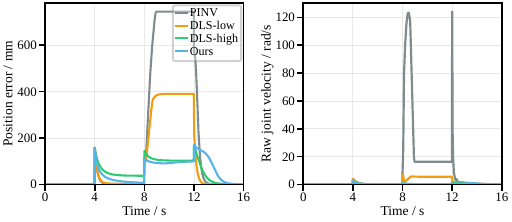}
  \caption{Singularity ablation. The proposed velocity mapping suppresses raw joint-velocity amplification while maintaining bounded position error near the workspace boundary.}
  \label{fig:singularity_ablation}
\end{figure}

\begin{table}[t]
\centering
\footnotesize
\setlength{\tabcolsep}{2.5pt}
\caption{Singularity ablation on reachable/unreachable tracking and velocity demand.}
\label{tab:singularity_ablation}
\begin{tabular}{@{}lrrrr@{}}
\toprule
\textbf{Method} & \textbf{Reachable}$\downarrow$ & \textbf{Unreachable}$\downarrow$ & \textbf{Min.\ cond.}$\uparrow$ & \textbf{Max.\ $\dot q$}$\downarrow$ \\
\midrule
PINV     & 0.0  & 745.2 & 0.0030 & 124.23 \\
DLS-low  & 1.5  & 390.3 & 0.0030 & 6.66 \\
DLS-high & 18.7 & 101.9 & 0.0267 & 1.52 \\
Ours     & \textbf{4.5} & \textbf{97.8} & 0.0030 & \textbf{2.63} \\
\bottomrule
\end{tabular}
\end{table}

\noindent\textbf{Singularity-robust velocity mapping.}
We next compare PINV, low- and high-damping least squares, and the proposed J-PARSE mapping on a home--boundary--unreachable--return sequence. Figure~\ref{fig:singularity_ablation} and Table~\ref{tab:singularity_ablation} show that PINV produces severe velocity amplification near the unreachable target, whereas Ours limits the peak request to $2.63$\,rad/s with a $97.8$\,mm residual. In this test, the mapping balances reachable-target accuracy and peak raw joint-velocity demand.

\section{Conclusion}
This work presents a human-like redundancy-resolution framework for real-time bimanual humanoid teleoperation. SWGR maps XR upper-body motion to robot-scale wrist targets, AEAC converts observable human elbow geometry into an adaptive null-space preference, and a singularity-robust mapping stabilizes the primary wrist task. Experiments on a JAKA~K1 across static poses, dynamic trajectories, and bimanual tasks show that the framework improves elbow consistency on redundancy-sensitive motions and reduces abrupt joint motion near singular configurations. One limitation is that each arm is solved independently, which may limit coordinated bimanual accuracy. Future work will investigate predictive singularity avoidance, inter-arm coordination, and whole-body extensions.

\end{document}